\documentclass[conference]{IEEEtran}
\usepackage{amsmath,amsfonts}
\usepackage{graphicx}
\usepackage{url}
\usepackage{cite}
\usepackage{float}
\usepackage{placeins}

\title{A Convolution-Based Gait Asymmetry Metric for Inter-Limb Synergistic Coordination}

\author{\IEEEauthorblockN{Go Fukino}\and{Kanta Tachibana}}

\begin{document}

\maketitle

\begin{abstract}
This study focuses on the velocity patterns of various body parts during walking and proposes a method for evaluating gait symmetry. Traditional motion analysis studies have assessed gait symmetry based on differences in electromyographic (EMG) signals or acceleration between the left and right sides. In contrast, this paper models intersegmental coordination using an LTI system and proposes a dissimilarity metric to evaluate symmetry. The method was tested on five subjects with both symmetric and asymmetric gait.
\end{abstract}

\begin{IEEEkeywords}
Gait analysis, convolution, transfer function, time-series, symmetry, OpenPose.
\end{IEEEkeywords}

\section{Introduction}
% --- Replace with full introduction content ---
Since the advent of Homo sapiens, bipedal
locomotion has been a fundamental human movement,
characterized by alternating cyclic motion. This study
focuses on the velocity patterns of various body parts during
walking and explores a method for evaluating gait symmetry.
Previous studies evaluating gait symmetry have
primarily relied on metrics such as maximum, minimum, and
mean values or have visually assessed time-series patterns
through graphical representations. For instance, Wang et al.\cite{7852992}
and Li et al.\cite{7852993}   assessed asymmetry in knee flexion and
extension during post-stroke gait using motion data and
electromyography (EMG), focusing on the maximum and
minimum knee joint angles and lower limb EMG signals.
Rathore et al.\cite{10103790}  investigated asymmetrical gait in unilateral
amputees using prosthetic limbs by employing a knee flexion
angle potentiometer and a foot pressure sensor to analyze
peak knee flexion angles and ground reaction forces. D'Arco
et al.\cite{10602869} evaluated gait symmetry based on plantar pressure
parameters obtained from smart insoles. Similarly, Loiret et
al.\cite{loiret2019wearable} utilized foot pressure sensors, as in Rathore et al. [3],
but specifically examined the peak values of ground reaction
forces at the thigh and hip level to assess symmetry.
Willamson et al.\cite{7299355} used accelerometers to evaluate
symmetry based on the mean acceleration values. While these
studies [1-5] incorporate temporal parameters such as stance
duration and gait cycle time, they primarily focus on
maximum, minimum, and average values of physical
quantities. However, they do not perform detailed analyses of
time-series patterns.
The studies conducted by Qin et al. and Yan et al.\cite{10123365,10716766,10663376,10458920}
 analyzed gait symmetry by plotting simultaneously
measured left and right electrostatic potential samples—
generated through friction between the body, the floor, or
clothing—on a plane and applying principal component
analysis (PCA). However, these studies [7]-[10] do not take
into account the sequential nature of the time-series data (i.e.,
that sample $n$ is followed by sample $n$ + 1) or the nonlinear
characteristics of time-series patterns.
Several studies have investigated gait symmetry through
detailed analysis of time-series patterns. Yogev et al. \cite{yogev2007gait}
evaluated gait symmetry in patients with Parkinson’s disease
and elderly individuals at risk of falling using a foot pressure
sensor system, considering swing time (the duration a foot
remains off the ground during walking). Adamczyk et al. \cite{6897983}
estimated velocity from ground reaction forces and assessed
gait symmetry through autocorrelation. Anna et al. \cite{5443473} and
Zhang et al. \cite{zhang2018gait} employed wearable accelerometers to
measure acceleration in both thighs and lower legs, using
cross-correlation to evaluate symmetry. Gouwanda et al. \cite{6498167}
and Sheng et al. \cite{8834082} proposed a symmetry evaluation method
based on cross-correlation of angular velocity data obtained
from wireless gyroscopes. Khoo et al. \cite{8626691} assessed gait
symmetry using time-series data of knee joint angles and joint
moments. Arauz et al. \cite{arauz2022spine} utilized motion capture
technology to analyze changes in joint angles during
treadmill walking (both normal-speed and high-speed
conditions) to evaluate gait symmetry. Lena et al. \cite{siebers2021comparison}
proposed a symmetry index by integrating a symmetry
function derived from time-series data of foot joint angles
over the gait cycle. Additionally, Diaz et al. \cite{ochoa-diaz2020symmetry} analyzed the
vertical trajectory of the body’s center of mass in lower-limb 
amputees (particularly transfemoral amputees) and
introduced a frequency-domain gait symmetry evaluation
method using Discrete Fourier Transform (DFT). These
studies have compared left and right time-series patterns in
both the time and frequency domains as symmetry evaluation
metrics, employing analytical approaches similar to
Proposed Method I in this study.
None of the existing methods evaluate the left-right
difference in Time Series Pairs (TSPs), as proposed in
Proposed Method II of this study. Additionally, these
conventional studies rely on specialized equipment, making
them less accessible for general use. This study aims to
develop a simple and practical gait symmetry evaluation
system that does not require dedicated devices, utilizing
smartphone video recordings and OpenPose. OpenPose
extracts skeletal information from video footage captured by
a standard camera. Since no specialized equipment is needed,
this approach significantly reduces the burden of both
preparation and measurement, making gait symmetry
analysis more accessible and convenient.
The objective of this study is to propose a
dissimilarity metric for evaluating the left-side and rightside intersegmental coordination systems underlying Time
Series Pairs (TSPs) extracted from video recordings of gait,
without the need for specialized equipment. Furthermore, this
study aims to determine whether the proposed metric can
effectively quantify left-right asymmetry in gait patterns.

\section{Theoretical Background}
In this study, we propose two indices of left-right
symmetry for analyzing cyclic alternating movements such
as walking. Proposed Method I: A method that compares left
and right movement patterns by shifting half a walking cycle;
Proposed Method II: A method that compares coordinated
systems based on transfer functions.
Proposed Method I measures waveform similarity
by shifting the movement velocity calculated from the left
and right body coordinates by +1/4 cycle for the left side and
-1/4 cycle for the right side. Proposed Method II uses the
transfer function between time-series signals of two body
parts. Specifically, it evaluates left-right symmetry using a
non-similarity index based on an LTI (Linear Time Invariant)
system, where the movement of the wrist (horizontal or
vertical) is analyzed in relation to the movement of the ankle
(horizontal or vertical).

\subsection{1/4 Cycle Shift Transformation}
For a discrete-time series of period \( N \), denoted as
\[
x(n), \quad n = 0, 1, \ldots, N-1,
\]
the Fourier coefficient \( X(k) \in \mathbb{C} \) of waveform \( k \) is given by:
\[
X(k) = \sum_{n=0}^{N-1} x(n) \exp\left(-j \frac{2\pi k n}{N}\right), \quad k = -K, \ldots, -1, 0, 1, \ldots,K,
\]
where \( j \) is the imaginary unit and \( K = \left\lfloor \frac{N}{2} \right\rfloor \).

For a periodic function \( x(t) \) with period \( T \), the \( \frac{1}{4} \) cycle shift transformation is represented as:
\[
x\left(t - \frac{T}{4}\right) = \sum_{k=-\infty}^{\infty} X(k) \exp\left(j \frac{2\pi k}{T} \left(t - \frac{T}{4}\right)\right).
\]
This becomes:
\[
= \sum_{k=-\infty}^{\infty} X(k) \exp\left(j \frac{2\pi k t}{T} \right) \exp\left(-j \frac{\pi k}{2}\right),
\]
\[
= \sum_{k=-\infty}^{\infty} (-j)^k X(k) \exp\left(j \frac{2\pi k t}{T}\right).
\]

This means that multiplying each Fourier coefficient \( X(k) \) by \( -j \) results in a \( +\frac{1}{4} \) cycle shift, while multiplying by \( j \) results in a \( -\frac{1}{4} \) cycle shift.

For observed discrete-time right-side and left-side data \( x(n) \) and \( y(n) \), respectively, with sample size \( M \) (including 3–6 strides), this transformation shifts the left side by \( +\frac{1}{4} \) cycle and the right side by \( -\frac{1}{4} \) cycle, resulting in a total shift of \( \frac{1}{2} \) cycle.

\section*{Symmetry Evaluation}

The left-right movement symmetry is then evaluated using the correlation coefficient:
\[
\rho = \frac{\mathrm{cov}(x, y)}{\sigma_x \sigma_y},
\]
where:
\[
\bar{x} = \frac{1}{M} \sum_{n=0}^{M-1} x[n], \quad \bar{y} = \frac{1}{M} \sum_{n=0}^{M-1} y[n],
\]
\[
\sigma_x = \sqrt{\frac{1}{M} \sum_{n=0}^{M-1} (x[n] - \bar{x})^2}, \quad \sigma_y = \sqrt{\frac{1}{M} \sum_{n=0}^{M-1} (y[n] - \bar{y})^2}.
\]

\subsection{Transfer Function-Based Symmetry}
Let the relationship between input and output Let the relationship between input and output discrete-time signals \( x[n] \) and \( y[n] \), for \( n = 0, \ldots, N-1 \), be modeled as a Linear Time-Invariant (LTI) system. The sequences are transformed using the Z-transform:

\[
X(z) = x[0]z^0 + x[1]z^{-1} + \cdots + x[N-1]z^{-N+1},
\]
\[
Y(z) = y[0]z^0 + y[1]z^{-1} + \cdots + y[N-1]z^{-N+1}.
\]

The system's transfer function is defined as:
\[
G(z) = \frac{Y(z)}{X(z)}.
\]

To evaluate movement symmetry, two symmetrical pairs of body parts are selected. Let \( G_1(z) = \frac{B(z)}{A(z)} \) be the transfer function for the right-side system, and \( G_2(z) = \frac{Y(z)}{X(z)} \) for the left-side system. The movement is considered symmetric if the transfer functions are approximately equal:

\[
A(z)Y(z) \approx X(z)B(z).
\]

Assume:
- \( a[n] \): horizontal/vertical speed of the right ankle,
- \( b[n] \): speed of the right wrist,
- \( x[n] \): speed of the left ankle,
- \( y[n] \): speed of the left wrist.

Then the Z-domain products are:
\[
A(z)Y(z) = \left( \sum_{n=0}^{N-1} a[n] z^{-n} \right) \left( \sum_{n=0}^{N-1} y[n] z^{-n} \right) = \sum_{k=0}^{2N-2} (a * y)[k] z^{-k},
\]
\[
X(z)B(z) = \left( \sum_{n=0}^{N-1} x[n] z^{-n} \right) \left( \sum_{n=0}^{N-1} b[n] z^{-n} \right) = \sum_{k=0}^{2N-2} (x * b)[k] z^{-k},
\]
where \( * \) denotes linear convolution, and:
\[
(a * y)[k] = \sum_{l=0}^{k} a[l] y[k - l], \quad (x * b)[k] = \sum_{l=0}^{k} x[l] b[k - l],
\]
assuming \( a[n] = b[n] = x[n] = y[n] = 0 \) for all \( n \notin [0, N) \).

Let:
\[
\mathbf{u} = a * y, \quad \mathbf{v} = x * b,
\]
where \( \mathbf{u}, \mathbf{v} \in \mathbb{R}^{2N - 1} \).

\section*{Dissimilarity Measure for Symmetry}

To quantify the symmetry of movement, the dissimilarity between the two LTI systems is defined as:

\[
\text{Dis}((a,b),(x,y)) = \frac{\|\mathbf{u} - \mathbf{v}\|^2}{\|\mathbf{u}\| \cdot \|\mathbf{v}\|},
\]

where \( \|\cdot\| \) denotes the Euclidean norm. A smaller dissimilarity implies more symmetric movement between the left and right side systems.

This measure is used to evaluate movement symmetry. Fig. 1 illustrates an example of dissimilarity evaluation between LTI systems. 
\begin{figure}[htbp]
  \centering
  \includegraphics[width=0.5\textwidth]{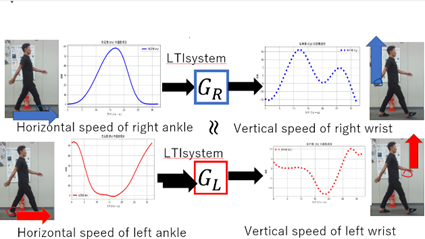}
  \caption{ Illustration of dissimilarity metric between LTI systems }
\end{figure}

\section{Methodology and Experiments}
We recorded the normal walking motion (referred to as Motion 
S) of five healthy male subjects. Immediately after, we asked them 
to perform an extremely asymmetrical walking motion (referred to 
as Motion A) and recorded it as well. Both motions were filmed 
from a lateral view at a distance of approximately 2 meters using a 
camera. The camera used was an LED light web camera (model NB
05) with a frame rate of 30 fps. 
OpenPose was used to extract joint coordinates for both 
Motion S and Motion A. The tracked joints included both 
ankles, both knees, both hips, both wrists, and both shoulders. 
Simultaneously, the confidence score for each joint 
coordinate estimation was recorded for each frame. If a joint 
was occluded or difficult to estimate, its confidence score 
dropped below 0.5, and its coordinates were not recorded. In 
cases where the confidence score was 0.5 or lower, linear 
interpolation was applied using the coordinates from adjacent 
frames. Next, a moving average filter with a window size of 
3 was applied to smooth the data. 
Figure 2 illustrates a single frame of normal walking, 
while Figure 3 shows a single frame of asymmetric walking. 
Table 1 summarizes the stride count, the number of frames in 
which the entire body of the subject is visible, and the gait 
cycle (in frames) computed via autocorrelation analysis for 
both symmetric and asymmetric walking motions for each 
subject.

\begin{table}[h!]
\centering
\caption{Recoded number of strides, frames and walking cycle in symmetric and asymmetric gait motion}
\vspace{1em}

% -------- Normal Gait Motion Table --------
\textbf{symmetri Gait Motion} \\
\vspace{0.5em}
\begin{tabular}{|c|ccc|}
\hline
\multicolumn{1}{|c|}{} & \multicolumn{3}{c|}{\textbf{motion S}} \\
\cline{2-4}
\textbf{subject} & stride & frames & cycle \\
\hline
1 & 3 & 80  & 33 \\
2 & 3 & 75  & 27 \\
3 & 3 & 60  & 35 \\
4 & 3 & 100 & 45 \\
5 & 4 & 100 & 33 \\
\hline
\end{tabular}

\vspace{2em}

% -------- Asymmetric Walk Table --------
\textbf{Asymmetric Walk} \\
\vspace{0.5em}
\begin{tabular}{|c|ccc|}
\hline
\multicolumn{1}{|c|}{} & \multicolumn{3}{c|}{\textbf{motion S}} \\
\cline{2-4}
\textbf{subject} & stride & frames & cycle \\
\hline
1 & 7 & 150 & 34 \\
2 & 3 & 75  & 35 \\
3 & 4 & 160 & 41 \\
4 & 6 & 150 & 66 \\
5 & 4 & 105 & 31 \\
\hline
\end{tabular}

\end{table}

\vspace{2em}

\subsection{Left-Right Symmetry Index Based on 1/4 Shift Transformation}

The movement velocity of each joint within a frame is calculated using the following formula:

\begin{equation}
v = \Delta x + \Delta y
\end{equation}

where $\Delta x$ and $\Delta y$ represent the differences in horizontal and vertical coordinates between consecutive frames, respectively. The velocity is expressed in units of [Pixels/Frame].

For the left and right ankle speed time series, autocorrelation analysis is performed to determine the gait cycle. Then, a $\frac{1}{4}$ shift transformation is applied: shifting the left ankle speed by $+\frac{1}{4}$ cycle and the right ankle speed by $-\frac{1}{4}$ cycle. This transformation is used to evaluate the left-right symmetry of ankle movement.

\begin{figure}[htbp]
  \centering
  \includegraphics[width=0.27\textwidth]{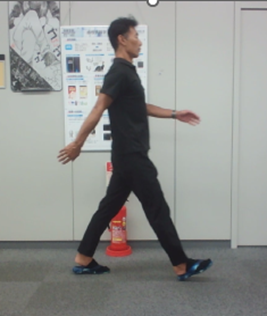}
  \caption{ Illustration of dissimilarity metric between LTI systems }
\end{figure}

\begin{figure}[htbp]
  \centering
  \includegraphics[width=0.25\textwidth]{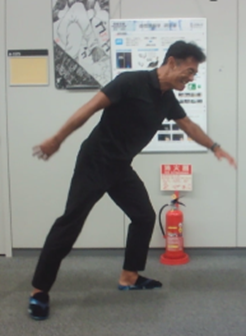}
  \caption{ Illustration of dissimilarity metric between LTI systems }
\end{figure}

\subsection{Symmetry Index Based on Transfer Function}
In the approach using transfer functions, the left-right 
symmetry of movement data is evaluated based on the 
horizontal or vertical motion of the ankles and wrists. This 
method assesses whether the movements exhibit symmetry 
between the left and right sides. 
Figure 4 summarizes the horizontal and vertical velocity 
components used to evaluate the symmetry between the 
ankles and wrists.

\begin{figure}[htbp]
  \centering
  \includegraphics[width=0.54\textwidth]{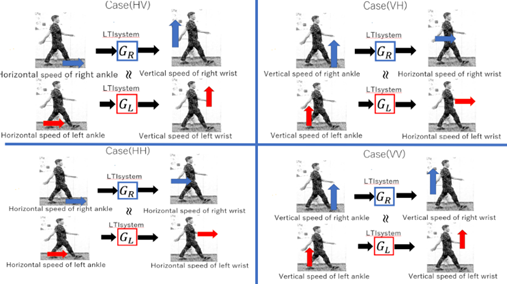}
  \caption{Illustration of symmetry evaluation using }
\end{figure}

horizontal and vertical ankle and wrist velocity components 
analyzed with LTI systems. Case (HV): The horizontal 
speed of the ankle is used as input, and the vertical speed of 
the wrist is output; Case (VH): The vertical speed of the 
ankle is used as input, and the horizontal speed of the wrist 
is output; Case (HH): The horizontal speed of the ankle is 
used as input, and the horizontal speed of the wrist is output;  
Case (VV): The vertical speed of the ankle is used as input, 
and the vertical speed of the wrist is output.

\section{Results}
\subsection{Symmetry Index Based on 1/4 Cycle Shift Transformation}
Fig. 5 shows the speed of the left and right ankles for the 
symmetric motion S, shifted by a total of half a cycle using a 
1/4 cycle shift transformation. Similarly, Fig. 6 illustrates the 
speed of the left and right ankles for the asymmetric motion 
A, also shifted by half a cycle. Table 2 presents the 
correlation results of the 1/4 cycle shift transformation for 
five individuals. 

\begin{figure}[htbp]
  \centering
  \includegraphics[width=0.5\textwidth]{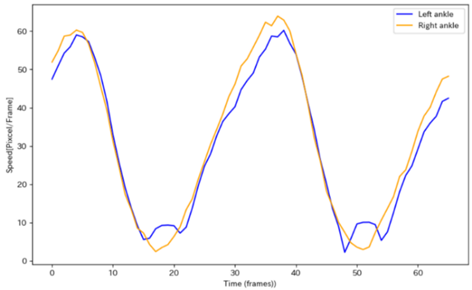}
  \caption{ Left and right ankle speed in motion S after 1/4 period shift transformation }
\end{figure}

\begin{figure}[htbp]
  \centering
  \includegraphics[width=0.5\textwidth]{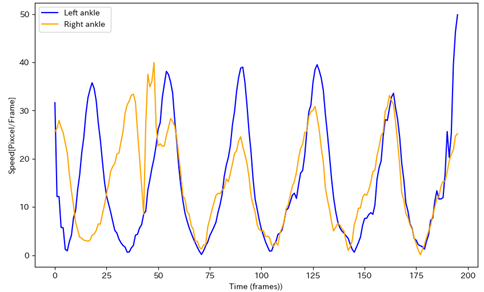}
  \caption{ Left and right ankle speed in motion A after 1/4 period shift transformation }
\end{figure}

\begin{table}[h]
\centering
\caption{ Correlation coefficients between left and right ankle speed after 1/4 period shift transformation }
\begin{tabular}{|c|c|c|}
\hline
\textbf{subject} & \textbf{Symmetric motion S} & \textbf{Asymmetric motion A} \\
\hline
1 & 0.98 & 0.46 \\
2 & 0.98 & 0.42 \\
3 & 0.92 & 0.70 \\
4 & 0.95 & 0.80 \\
5 & 0.98 & 0.64 \\
\hline
\end{tabular}

\label{tab:motion_scores}
\end{table}

\subsection{SYMMETRY INDEX BASED ON TRANSFER FUNCTION }

Table 3 presents the asymmetry evaluation results for five 
pedestrians. The (S) column represents the dissimilarity for 
the symmetric motion S, while the (A) column represents the 
dissimilarity for motion A. 
For motion S, the dissimilarity was less than 1 for all five 
individuals across all four cases, indicating a high level of 
symmetry. In contrast, for motion A, the dissimilarity values 
were 1 or greater in cases (HV), (VH), and (VV) for all five 
individuals, indicating low symmetry. However, in case 
(HH), the dissimilarity remained below 1 for all five 
individuals. 
Fig. 7 illustrates the horizontal and vertical speed of 
subject 1 during the symmetric motion S. The gait cycle was 
determined through autocorrelation analysis, and the data 
were segmented and overlaid based on the gait cycle. The 
horizontal axis represents frames, while the vertical axis

epresents horizontal speed components. The different 
colors indicate: 
\begin{itemize}
  \item \textbf{Blue}: Right ankle
  \item \textbf{Red}: Right wrist
  \item \textbf{Green}: Left ankle
  \item \textbf{Yellow}: Left wrist
\end{itemize}

Similarly, Fig. 8 shows the horizontal and vertical speed 
of subject 1 during the asymmetric motion A. 
Fig. 9 presents the convolution diagram for case (HH) of 
the symmetric motion S, while Fig. 10 shows the convolution 
diagram for case (HH) of the asymmetric motion A.

\begin{table}[h]
\centering
\caption{Summary of dissimilarity value\\
$\mathrm{Dis}((\text{Right ankle}, \text{Right wrist}), (\text{Left ankle}, \text{Left wrist}))$}
\label{tab:dissimilarity_summary}
\begin{tabular}{|c|cc|cc|cc|cc|}
\hline
\textbf{subject} 
& \multicolumn{2}{c|}{\textbf{case(HV)}} 
& \multicolumn{2}{c|}{\textbf{case(VH)}} 
& \multicolumn{2}{c|}{\textbf{case(HH)}} 
& \multicolumn{2}{c|}{\textbf{case(VV)}} \\
\cline{2-9}
 & (S) & (A) & (S) & (A) & (S) & (A) & (S) & (A) \\
\hline
1 & 0.32 & 1.88 & 0.49 & 1.40  &   0.20    &  0.22     &   0.31    &    1.81   \\
2 & 0.62 & 2.97 & 0.39 & 1.60  &   0.29    &  0.43    &     0.92  &     4.40  \\
3 & 0.41 & 2.00 & 0.17 & 0.78 &    0.15  &     0.79  &     0.18  &      0.80 \\
4 & 0.64 & 2.00 & 0.57 & 1.94 &     0.44  &    0.52   &    0.80   &     2.13  \\
5 & 0.29 & 1.57 & 0.19 & 0.27 &     0.20  &    0.09   &    0.25   &      2.00 \\
\hline
\end{tabular}
\end{table}

\begin{figure}[H]
  \centering
  \includegraphics[width=0.5\textwidth]{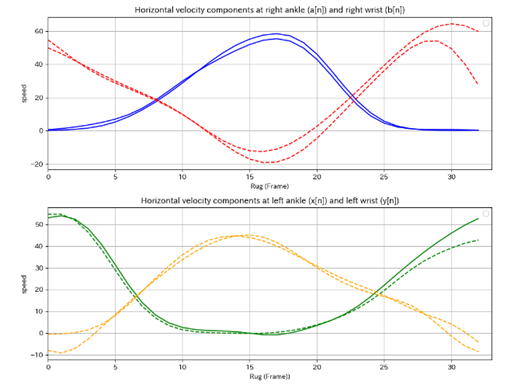}
  \caption{Horizontal speed in motion S of subject 1. Blue: Right Ankle; Red: Right Wrist; Green: Left Ankle; Yellow: Left Wrist}
\end{figure}

\begin{figure}[H]
  \centering
  \includegraphics[width=0.5\textwidth]{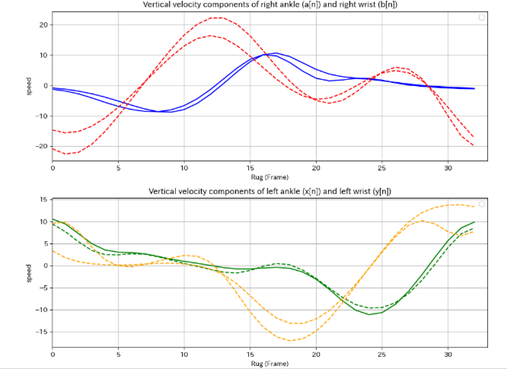}
  \caption{Vertical speed in motion S of subject 1. Blue: Right Ankle; Red: Right Wrist; Green: Left Ankle; Yellow: Left Wrist}
\end{figure}

\begin{figure}[H]
  \centering
  \includegraphics[width=0.5\textwidth]{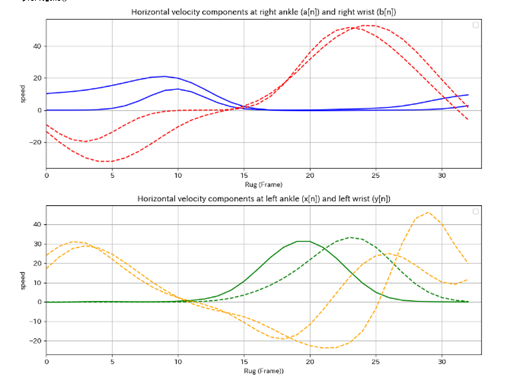}
  \caption{Horizontal speed in motion A of subject 1. Blue: Right Ankle; Red: Right Wrist; Green: Left Ankle; Yellow: Left Wrist}
\end{figure}
\FloatBarrier % subsectionの直前に入れることで、前の図が越えないようにする
\begin{figure}[H]
  \centering
  \includegraphics[width=0.5\textwidth]{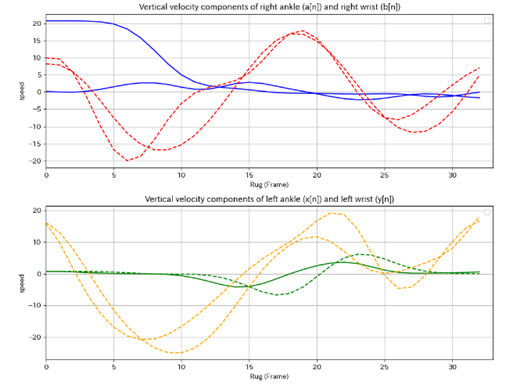}
  \caption{Vertical speed in motion A of subject 1. Blue: Right Ankle; Red: Right Wrist; Green: Left Ankle; Yellow: Left Wrist}
\end{figure}

\begin{figure}[H]
  \centering
  \includegraphics[width=0.5\textwidth]{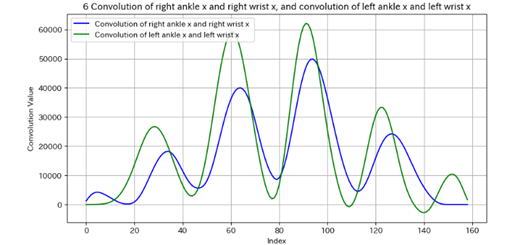}
  \caption{Convolution computed case (HH) for motion S of subject 1}
\end{figure}

\begin{figure}[H]
  \centering
  \includegraphics[width=0.5\textwidth]{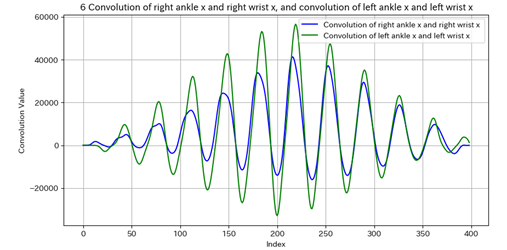}
  \caption{Convolution computed case (HH) for motion A of subject 1}
\end{figure}

\FloatBarrier % この図たちをまとめてsubsection内に閉じ込めたいなら、最後にも入れる

\section{Discussion}

The effectiveness of the symmetry index, which shifts the motion by a total of half a cycle using a \( \frac{1}{4} \) shift transformation for alternating periodic movements, was verified for the symmetric motion (\textbf{S}) and asymmetric motion (\textbf{A}) in five individuals.

When evaluating the dissimilarity of the transfer function from ankle movement to the movement of the ipsilateral wrist, excluding case (\textbf{HH}), the asymmetric movement \textbf{A} showed a higher value, indicating greater asymmetry compared to the symmetric movement \textbf{S}. Clearly, asymmetric movements exhibited a high degree of dissimilarity in the left and right transfer functions. On the other hand, in asymmetric performances, all five participants showed similar left and right transfer functions for the horizontal movement of the ankle to the horizontal movement of the ipsilateral wrist in case (\textbf{HH}).

Figure~11 presents a box plot of the dissimilarity values for the five participants. The reason for the low dissimilarity in case (\textbf{HHA}) is that, while asymmetric movement \textbf{A} is a performance where intentional postures and movements can be expressed, the dynamic characteristics of the participants' motor coordination are instinctive, making it difficult to completely conceal their inherent left–right symmetry.

In this study, the input and output of the LTI system were represented as follows: \( \vec{a}[n] \cdot \vec{e}_1 \) and \( \vec{b}[n] \cdot \vec{e}_1 \), where the basis vectors \( \vec{e}_1 \) and \( \vec{e}_2 \) correspond to the horizontal and vertical directions, respectively. In the future, we aim to extend this approach by considering an LTI system that directly processes the two-dimensional vectors \( \vec{a}[n] \) and \( \vec{b}[n] \) as inputs and outputs.

In this extended framework, each convolution term \( \vec{a}[n]\,\vec{y}[k - n] \) would not be a real number but rather the geometric product of vectors. Consequently, the dissimilarity measure would be defined as the ratio of the norms of complex numbers, ensuring a non-negative real value. A similar formulation can be derived for three-dimensional vectors, where the dissimilarity measure is defined as the ratio of the norms of quaternions.

\begin{figure}[H]
  \centering
  \includegraphics[width=0.5\textwidth]{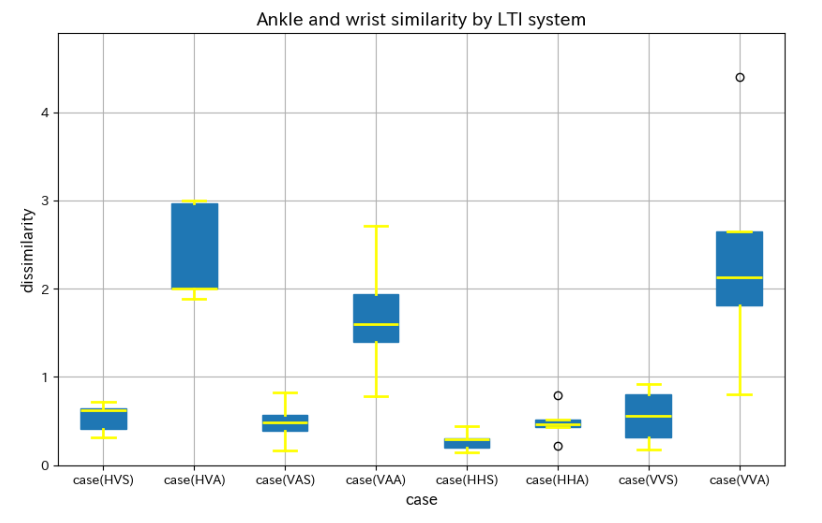}
  \caption{ Box-and-whisker diagram of dissimilarities evaluated for each case of 5 subjects}
\end{figure}

\section{Conclusion}
In this study, we propose a dissimilarity measure for the
left-side and right-side coupling systems underlying the time
series pair (TSP) of two body points during walking, using
video captured by a camera without the need for specialized
equipment. We then examined whether the proposed
measure can quantitatively assess left-right differences.
When evaluating the left-right difference in transfer
functions using an LTI system, the transfer function from
the horizontal movement of the ankle to the ipsilateral wrist
in asymmetric movements showed similarity between the
left and right sides, revealing the pedestrian's inherent leftright symmetry.
All data and source code from this study are available at

\bibliographystyle{IEEEtran}

\bibliography{reference}

\end{document}